\documentclass[letterpaper, 10 pt, conference]{ieeeconf}  % Comment this line out if you need a4paper

\usepackage{algorithm}
\usepackage[noend]{algpseudocode}
\usepackage{amsfonts}
\usepackage{amssymb}
\usepackage{bm}
\usepackage{graphicx}
\usepackage{multirow}
\usepackage{subcaption}
\usepackage{hyperref}
\usepackage{moreverb,url}

{
\addtocounter{footnote}{1} % to get dagger instead of star
\renewcommand\thefootnote{\@fnsymbol\c@footnote}%
}

\IEEEoverridecommandlockouts                              % This command is only needed if 
\title{\LARGE \bf
Learning One-Shot Imitation from Humans without Humans
}

\author{Alessandro Bonardi$^{1}$, Stephen James$^{2}$, Andrew J. Davison$^{2}$% <-this % stops a space
\thanks{$^{1}$Department of Computing, Imperial College London}%
\thanks{$^{2}$Dyson Robotics Lab, Imperial College London}%
\thanks{Research presented here has been supported by Dyson Technology Ltd}% <-this % stops a space
}

\renewcommand{\thefootnote}{\fnsymbol{footnote}}

\newcommand{\policy}{\ensuremath{\pi}}
\newcommand{\expertpolicy}{\ensuremath{\pi^*}}
\newcommand{\action}{\ensuremath{\boldsymbol{a}}}

\newcommand{\obs}{\ensuremath{\boldsymbol{o}}}
\newcommand{\task}{\ensuremath{\mathfrak{T}}}
\newcommand{\example}{\ensuremath{\tau}}

\newcommand{\sentence}{\ensuremath{\boldsymbol{s}}}
\newcommand{\emb}{\ensuremath{f_\theta}}
\newcommand{\loss}{\ensuremath{\mathcal{L}}}
\newcommand{\margin}{\ensuremath{margin}}
\newcommand{\support}{\ensuremath{U}}
\newcommand{\query}{\ensuremath{Q}}

\newcommand{\robotcollection}{\ensuremath{\mathfrak{R}}}
\newcommand{\humancollection}{\ensuremath{\mathfrak{H}}}
\newcommand{\tecnets}{TecNets}

\begin{document}

\maketitle
\thispagestyle{empty}
\pagestyle{empty}

%%%%%%%%%%%%%%%%%%%%%%%%%%%%%%%%%%%%%%%%%%%%%%%%%%%%%%%%%%%%%%%%%%%%%%%%%%%%%%%%
\begin{abstract}
Humans can naturally learn to execute a new task by seeing it performed by other individuals once, and then reproduce it in a variety of configurations. Endowing robots with this ability of imitating humans from third person is a very immediate and natural way of teaching new tasks. Only recently, through meta-learning, there have been successful attempts to one-shot imitation learning from humans; however, these approaches require a lot of human resources to collect the data in the real world to train the robot. But is there a way to remove the need for real world human demonstrations during training? We show that with Task-Embedded Control Networks, we can infer control polices by embedding human demonstrations that can condition a control policy and achieve one-shot imitation learning. Importantly, we do not use a real human arm to supply demonstrations during training, but instead leverage domain randomisation in an application that has not been seen before: sim-to-real transfer on humans. Upon evaluating our approach on pushing and placing tasks in both simulation and in the real world, we show that in comparison to a system that was trained on real-world data we are able to achieve similar results by utilising only simulation data. Videos can be found here\footnote{\url{https://sites.google.com/view/tecnets-humans}}.
\end{abstract}

%%%%%%%%%%%%%%%%%%%%%%%%%%%%%%%%%%%%%%%%%%%%%%%%%%%%%%%%%%%%%%%%%%%%%%%%%%%%%%%%
\section{INTRODUCTION}

Humans are able to learn how to perform a task by simply observing their peers performing it once; this is a highly desirable behaviour for robots, as it would allow the next generation of robotic systems, even in households, to be easily taught tasks, without additional technology or long interaction times. Endowing a robot with the ability to learn from a single human demonstration rather than through teleoperation, would allow for a more seamless human-robot interaction. 

\begin{figure}
    \centering
    \includegraphics[width=0.9\linewidth]{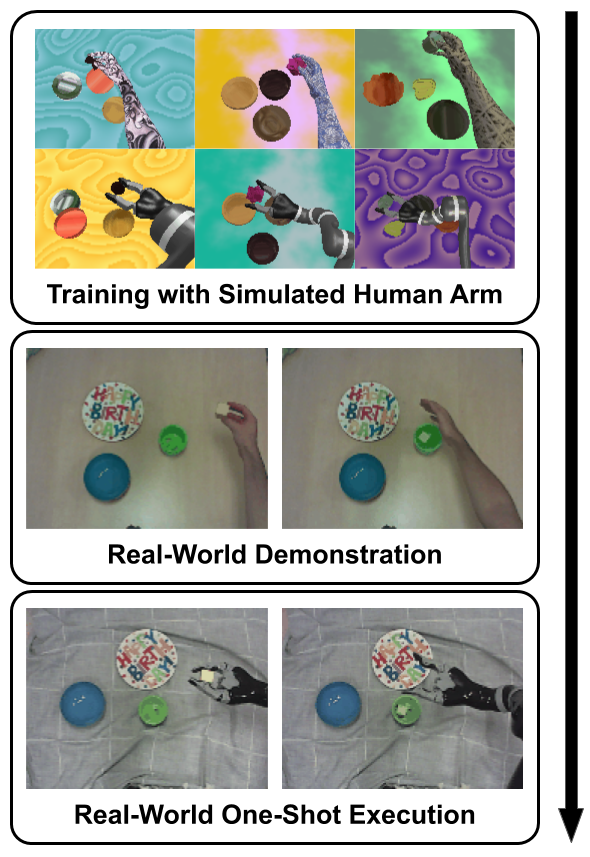}
    \caption{The robot gains its ability to infer actions from humans in simulation, and can then learn a new task from a single human demonstration in the real-world.}
    \label{fig:sequence}
\end{figure}

\begin{figure*}
    \centering
    \includegraphics[width=1.0\linewidth]{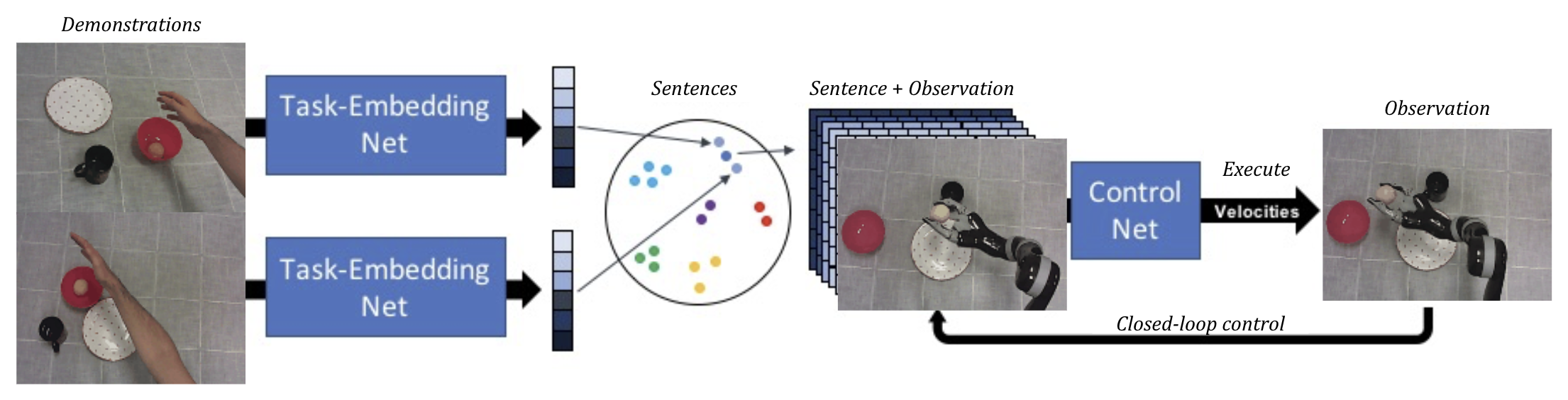}
    \caption{We use Task-Embedded Control Networks (\tecnets) to allow tasks to be learned from a single human demonstration. Images of human demonstrations are embedded into a compact representation of a task, which is then expanded and concatenated (channel-wise) to the most recent observation from a new configuration of that task before being sent through the control network in a closed-loop manner. Both the task-embedding net and control net are jointly optimised to produce a rich embedding.}
    \label{fig:approach}
\end{figure*}

Previous work has investigated hand-engineered systems which track movements and specify a mapping between the human and robot domains \cite{lee2013syntactic, yang2015robot}. Rather than explicitly hand-engineered systems, an emerging trend in robotics is to instead learn control directly from raw sensor data in an end-to-end manner. These systems operate well when close and complicated coordination is required between vision and control \cite{levine2016end}. Domain-Adaptive Meta-Learning (DAML) is a recent approach that uses an end-to-end method for one-shot imitation of humans \cite{yu2018one} which leveraged a large amount of prior meta-training data collected for many different tasks. This approach required thousands of examples across many tasks during meta-training: these examples are videos of a person physically performing the tasks and teleoperated robot demonstrations, meaning that there has to be an active and long human presence when collecting the dataset. Is there a way to reduce, or even eliminate, the amount of human presence that is needed when collecting datasets that require footage of humans? We propose that the recent successes in visual simulation-to-reality transfer \cite{james2017transferring, matas2018sim, bousmalis2018using, james2019sim} suggest there is a way.

To that end, we present an approach to the one-shot human imitation learning problem which does not require an active manual intervention during training, thus saving tens or hundreds of researchers hours. We show that the recent work on Task-Embedded Control Networks (\tecnets{}) \cite{james2018task} can be used to infer control polices by embedding human demonstrations that can condition a control policy and achieve one-shot imitation learning. Rather than using real humans to supply demonstrations during training, we instead leverage domain randomisation in an application that has not been seen before: sim-to-real transfer on humans. After training, we are able to deploy a system in the real world which can perform a previously unseen task in a new configuration after a single real-world human demonstration. Our approach, which is summarised in Figure \ref{fig:sequence}, is evaluated on pushing and placing tasks in both simulation and in the real world. We show we are able to achieve similar results to a system trained on real-world data. Moreover, we show that our approach remains robust to visual domain-shifts, such as a substantially different background, between the human demonstrator and the robot agent performing the task.

% For example, having the human supply a demonstration with a cloth on the table, and then getting the robot to perform the same task but with the table cloth removed.

% Performance does no degrade when there is a large domain shift between the demonstrtion and the 

\section{RELATED WORK}

\textbf{Imitation learning} aims to learn tasks by observing a demonstrator, and can broadly be classified into two  key areas: (1) \textit{behaviour cloning}, where an agent learns a mapping from observations to actions given demonstrations \cite{pomerleau1989alvinn, ross2011reduction}, and (2) \textit{inverse reinforcement learning} \cite{ng2000algorithms}, where an agent attempts to estimate a reward function that describes the given demonstrations \cite{abbeel2004apprenticeship, finn2016guided}. In this work we concentrate on the former. The majority of work in behaviour cloning operates on a set of configuration-space trajectories that can be collected via tele-operation \cite{calinon2009learning, zhang2017deep}, kinesthetic teaching \cite{akgun2012trajectories, pastor2011online}, sensors on a human demonstrator \cite{dillmann2004teaching, ekvall2004interactive, calinon2006teaching, kruger2010learning}, through motion planners \cite{james2017transferring}, or even by observing humans directly. Expanding further on the latter, learning by observing humans has previously been achieved through hand-designed mappings between human actions and robot actions \cite{lee2013syntactic, yang2015robot, rothfuss2018deep}, visual activity recognition and explicit handtracking \cite{lee2017learning, ramirez2017transferring}, and more recently by a system that infers actions from a single video of a human via an end-to-end trained system \cite{yu2018one}.

% \cite{kjellstrom2008visual, yang2015robot, lee2017learning, ramirez2017transferring, rothfuss2018deep, nguyen2018translating, yu2018one}

\textbf{One-shot and few-shot learning} is the paradigm of learning from a small number of examples at test time, and has been widely studied in the image recognition community \cite{vinyals2016matching, koch2015siamese, santoro2016meta, ravi2017optimization, triantafillou2017few, snell2017prototypical}. Our approach is based on \textit{James et al.}~\cite{james2018task} which comes under the domain of metric learning \cite{kulis2012metric, bellet2013survey}. There is an abundance of work in metric learning, including Matching Networks \cite{vinyals2016matching}, which use an attention mechanism over a learned embedding space to produce a weighted nearest neighbour classifier given labelled examples called a support set and unlabelled examples called a query set. Prototypical Networks \cite{snell2017prototypical} are similar, but differ in that they represent each class by the mean of its examples, the prototype, and use a squared Euclidean distance rather than the cosine distance. In the case of one-shot learning, matching networks and prototypical networks become equivalent. 

\textbf{Sim-to-real} methods attempt to address the apparent domain gap of both the visual and dynamics between simulation and the real-world, which reduces the need for expensive real-data collection. It has been shown that naively transferring skills between the two domains is not possible \cite{james20163d}, resulting in numerous attempts at transfer methods in both computer vision and robotics. Domain randomisation, which applies random textures, lighting, and camera position to the simulated scenes, has seen great success in numerous vision-based robotics applications \cite{sadeghi2016cad2rl, tobin2017domain, james2017transferring, matas2018sim, bousmalis2018using}. This method allows the algorithm operating on these randomised scenes to become invariant to domain differences that appear in the real world. Rather than directly operating on randomised images, RCAN \cite{james2019sim} is a recent approach that instead translate randomised rendered images into their equivalent non-randomised, canonical versions, producing superior results on a complex sim-to-real grasping task. Rather than operating on RGB images, other works have instead used depth images to cross the domain gap~\cite{viereck2017learning, gualtieri2016high}; however, in our tasks, the colour of an object is an important feature when inferring what object the robot needs to interact with, particularly when the geometry of the objects are very similar. In our work, we show that domain randomisation can be leveraged to transfer the ability to infer actions from human demonstrations.  

\section{BACKGROUND}

Our approach builds on Task-Embedded Control Networks (\tecnets{}), which we summarise in the following. The method is composed of a \textit{task-embedding network} and a \textit{control network} that are jointly trained to output actions (e.g. motor velocities) for a new variation of a task, given one or more demonstrations of it. Using these demonstrations, the task-embedding network has the responsibility of learning a compact representation of a task, which we call a \textit{sentence}. The control network then takes this static sentence along with current observations of the world to output actions on a variation of the same task. \tecnets{} do not have a strict restriction on the number of tasks that can be learned, and do not easily forget previously learned tasks during training, or after. The setup only expects the observations (\textit{e.g.} visual) from the demonstrator during test time, which makes it very applicable for learning from human demonstrations. 

Formally, a policy $\policy$ for task $\task$ maps observations $\obs$ to actions $\action$, and we assume to have expert policies $\expertpolicy$ for multiple different tasks. Corresponding example trajectories consist of a series of observations and actions: $\example = [(\obs_1, \action_1), \ldots, (\obs_T, \action_T)]$ and we define each task to be a set of such examples, $\task = \{ \example_1, \cdots, \example_K \}$. \tecnets{} aim to learn a universal policy $\policy(\obs, \sentence)$ that can be modulated by a sentence $\sentence$, where $\sentence$ is a learned description of a task $\task$. The resulting universal policy $\policy(\obs, \sentence)$ should emulate the expert policy $\expertpolicy$ for task $\task$.

For training, we sample two disjoint sets of examples for every task $\task^j$: a support set $\task_\support^j$ and a query set $\task_\query^j$.  The support set is used to compute a combined sentence $\sentence^j \in \mathbb{R}^N$ for the task, by taking the normalised mean of the sentence for each example:
\begin{equation}
\label{eq:sentence}
    \sentence^j = \bigg[ \frac{1}{|\task_\support^j|} \sum_{\example \in \task_\support^j} \emb(\example) \bigg] ^\wedge~,
\end{equation}
where $\boldsymbol{v}^\wedge = \frac{\boldsymbol{v}}{\|\boldsymbol{v}\|}$ and where $\emb$ is the embedding network. Using a combination of the cosine distance between points and the hinge rank loss (inspired by \cite{frome2013devise}), the loss for a query set  $\task_\query^j$ is defined as:
\begin{equation}
\label{eq:hinge_rank}
    \loss_{emb} = \sum_{\example \in \task_\query^j} \sum_{i \neq j} max[0, \margin - \emb(\example) \cdot \sentence^j + \emb(\example) \cdot \sentence^i]~.
\end{equation}

This loss helps learning an embedding space in which tasks that are visually and semantically similar are also close in the embedding space. Additionally, given a sentence $\sentence^j$, computed from the support set $\task_\support^j$, as well as examples from the query set $\task_\query^j$, the following behaviour-cloning loss for the policy $\policy$ can be computed:
\begin{equation}
    \loss_{ctr}^{\query} = \sum_{\example \in \task_\query^j} \sum_{(\obs, \action) \in \example} \| \policy(\obs, \sentence^j) - \action \|^2_2 ~.
\end{equation}

It was found that having the control network also predict the action for the examples in the support set $\task_\support^j$ leads to increased performance. Thus, the final loss is:
\begin{equation}
    \loss_{Tec} = \sum_{\task} \lambda_{emb} \loss_{emb} + \lambda_{ctr}^{\support} \loss_{ctr}^{\support} + \lambda_{ctr}^{\query} \loss_{ctr}^{\query}.
\end{equation}

\section{LEARNING FROM HUMANS USING TECNETS}

\begin{algorithm}[t!]
\centering
\caption{Training loss computation for one batch. $B$ is the batch size, $K_{\support}$ and $K_{\query}$ are the number of examples from the support and query set respectively, $K_{\robotcollection}$ is the number of robot examples to sample, and $RandomSample(S, N)$ selects $N$ elements uniformly at random from the set $S$.}
\label{algo:tecnets}
\begin{algorithmic}[1]
\Procedure{Training Iteration}{}
\State $\mathcal{B} = RandomSample(\{\task_1, \cdots, \task_N\}, B)$
\For{$\task^j \in \mathcal{B}$}
    \State $(\robotcollection^j, \humancollection^j) \gets \task^j$
	\State $\humancollection_\support^j = RandomSample(\humancollection^j, K_{\support})$
	\State $\humancollection_\query^j = RandomSample(\humancollection^j \backslash \humancollection_\support^j, K_{\query})$
% 	\State $\robotcollection_\support^j = RandomSample(\robotcollection^j, K_{\support})$
    \State $\robotcollection^j_{'} = RandomSample(\robotcollection^j, K_{\robotcollection})$
    
    \State $\sentence_\support^j = \Big[ \frac{1}{K_{\support}} \sum_{\example \in \humancollection_\support^j} \emb(\example) \Big] ^\wedge$
    \State $\sentence_q^j = \emb(\example_q) \quad \forall \example_q \in \humancollection_\query^j$
\EndFor
$\loss_{emb} = \loss_{\humancollection_{ctr}}^{\support} = \loss_{\robotcollection_{ctr}} = 0$
\For{$\task^j \in \mathcal{B}$}
	\State $\loss_{emb} \mathrel{+}= \sum_{q} \sum_{i \neq j} max[$ \parbox[t]{.38\linewidth}{%
	$0, \margin - \sentence_q^j \cdot \sentence_\support^j + \sentence_q^j \cdot \sentence_\support^i]$}
    \State $\loss_{\humancollection_{ctr}}^{\support} \mathrel{+}= \sum_{\example \in \humancollection_\support^j} \sum_{(\obs, \action) \in \example} \| \policy(\obs, \sentence_\support^j) - \action \|^2_2$
    \State $\loss_{\robotcollection_{ctr}} \mathrel{+}= \sum_{\example \in \robotcollection^j_{'}} \sum_{(\obs, \action) \in \example} \| \policy(\obs, \sentence_\support^j) - \action \|^2_2$
\EndFor
\State $\loss_{tec} = \lambda_{emb} \loss_{emb} + \lambda_{\humancollection_{ctr}}^{\support} \loss_{\humancollection_{ctr}}^{\support} + \lambda_{\robotcollection_{ctr}}^{\query} \loss_{\robotcollection_{ctr}}^{\query}$
\State \textbf{return} $\loss_{tec}$
\EndProcedure
\end{algorithmic}
\end{algorithm}

We expand on the \tecnets{} method introduced in the previous section by incorporating the notion of a human demonstrator which can be summarised in Figure \ref{fig:approach}. We slightly modify the definition of a task $\task$ to instead include two collections of examples: a human demonstrator collection $\humancollection = \{ \example_1, \cdots, \example_K \}$ and a robot agent collection $\robotcollection = \{ \example_1, \cdots, \example_K \}$, such that $\task = (\humancollection, \robotcollection)$. 
% Since we do not have access to the actions taken by the human demonstrator at test time, these are not included in the task and will also not be accessible to the task-embedding network.

From this, we now pick three disjoint sets of examples (rather than the original two) for every task $\task^j$: a support set of human examples $\humancollection_\support^j$, a query set of human examples $\humancollection_\query^j$, and a set of robot examples $\robotcollection^j_{'}$.

In analogy to Eq. (\ref{eq:sentence}) a combined sentence $\sentence^j \in \mathbb{R}^N$ for a task is computed by taking the normalised mean of the sentence for each example in the support set of human examples $\humancollection_\support^j$:
\begin{equation}
    \sentence^j = \bigg[ \frac{1}{|\humancollection_\support^j|} \sum_{\example \in \humancollection_\support^j} \emb(\example) \bigg] ^\wedge~,
\end{equation}

We then train the embedding model to produce a higher dot-product similarity between human demonstrations of a task's embedded example $\emb(\example)$ and its sentence $\sentence^j$ than to sentences of human demonstrations from other tasks $\sentence^i$:
\begin{equation}
    \loss_{emb} = \sum_{\example \in \humancollection^j} \sum_{i \neq j} max[0, \margin - \emb(\example) \cdot \sentence^j + \emb(\example) \cdot \sentence^i]~.
\end{equation}

Additionally, given a sentence $\sentence^j$, computed from the support set $\humancollection_\support^j$, as well as examples from the robot set $\robotcollection_{'}^j$ for the same task we can compute the following behaviour-cloning loss for the policy $\policy$:
\begin{equation}
    \loss_{ctr} = \sum_{\example \in \robotcollection_\query^j} \sum_{(\obs, \action) \in \example} \| \policy(\obs, \sentence^j) - \action \|^2_2 ~.
\end{equation}

The final loss is the combination of the embedding loss $\loss_{emb}$, the control loss on the support set for the human examples, and the control loss on the robot examples:
\begin{equation}
    \loss_{tec} = \lambda_{emb} \loss_{emb} + \lambda_{\humancollection_{ctr}}^{\support} \loss_{\humancollection_{ctr}}^{\support} + \lambda_{\robotcollection_{ctr}} \loss_{\robotcollection_{ctr}}.
\end{equation}
Note that only the human examples of the same task are explicitly enforced to be close together in the embedding space, rather than human and robot examples. Although we could have also enforced an additional embedding loss on human and robot examples being close together, in practice we found that this was not necessary. This is due to the joint training of both task-embedding and control networks which enforces the network to implicitly learn to map the embedded human examples to a set of corresponding robot actions. Pseudocode for both the training is provided in Algorithm \ref{algo:tecnets}.

Input to the task-embedding network consists of $(width, height, 3 \times |\example|)$, where $3$ represents the RGB channels. As in the \tecnets{} paper, we found that we only need to take the first and last frame of an example trajectory $\example$ for computing the task embedding and so we discard intermediate frames, resulting in an input of $(width, height, 6)$. The sentence from the task-embedding network is then tiled and concatenated channel-wise to the input of the control network (as shown in Figure \ref{fig:approach}), resulting in an input image of $(width \times height \times 3 + N)$, where $N$ represents the length of the embedding.

% TODO: mention that we dont take actions in the embedding space!

\subsection{Data Collection in Simulation} \label{datacollection}

\begin{figure}
    \centering
    \includegraphics[width=0.9\linewidth]{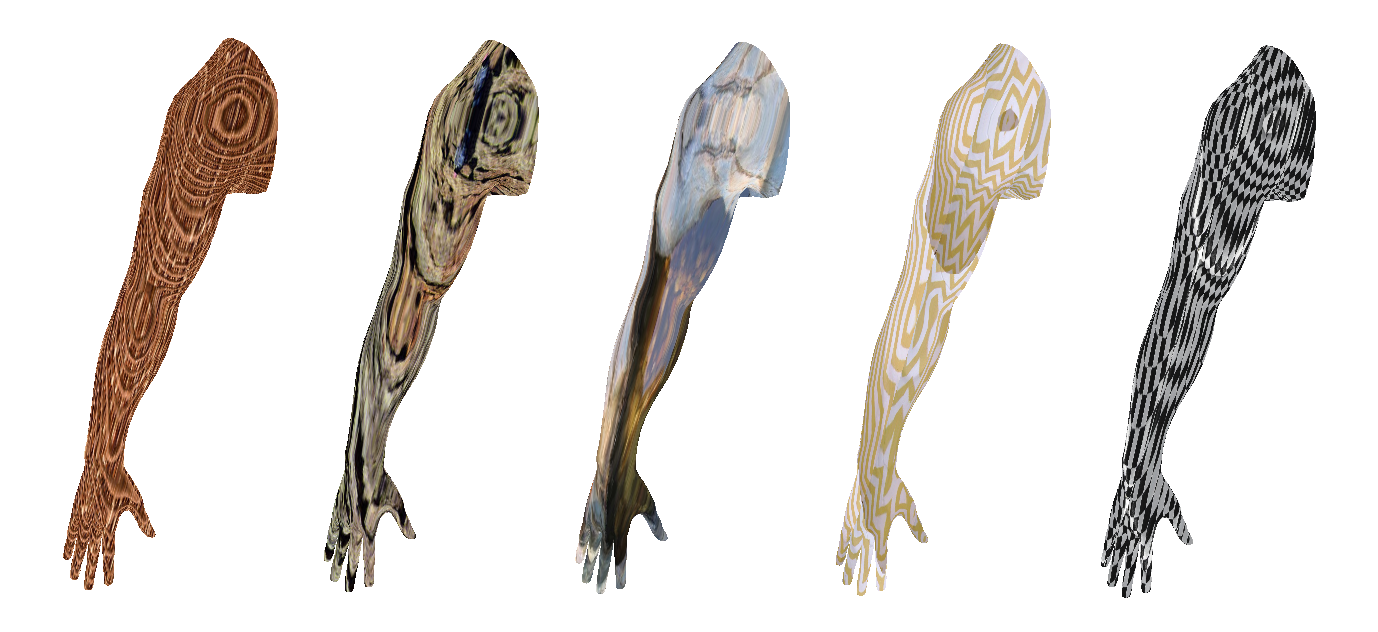}
    \caption{An example of the variations we get when we apply domain randomisation to the simulated human arm.}
    \label{fig:randomised_arm}
\end{figure}

Many approaches to human imitation rely on training in the real world. This has many disadvantages, but most evident is the amount of time and effort needed to collect data for the training dataset. In the case of DAML, thousands of demonstrations had to be recorded, which rely on an active human presence to obtain both human and robot demonstrations, as the robot still has to be controlled in some way. For instance, in the DAML placing experiment a total of $2586$ demonstrations were collected to form the training dataset, meaning tens of research hours dedicated to collecting data, with no guarantees that the dataset allows the network to generalise well enough. Training in simulation provides much more flexibility and availability of data: data generation can be easily parallelised and does not require constant human intervention. Additionally, there have been many successful examples of systems trained in simulation and then run in the real-word; one common approach to do this is domain randomisation \cite{sadeghi2016cad2rl, tobin2017domain, james2017transferring, matas2018sim, bousmalis2018using, james2019sim}.

Our approach generates the training dataset using PyRep \cite{james2019pyrep}, a recently released robot learning research toolkit, built on top of V-REP \cite{vrep}. We modelled a 3D mesh of a human arm from \href{https://www.nonecg.com/}{nonecg.com}, which we then broke down into rigid shapes. Our simulated arm has 26 degrees of freedom: 3 in the shoulder, 2 in the elbow, 2 for the wrist and the remaining 19 in the hand. 26 revolute joints link together the different rigid shapes: to emulate the soft-body behaviour of a real arm during motion, adjacent shapes slide over each other, making previously hidden parts of each shape visible. The resulting effect is very similar to real human arm motion.

\begin{figure}
    \centering
    \includegraphics[width=0.9\linewidth]{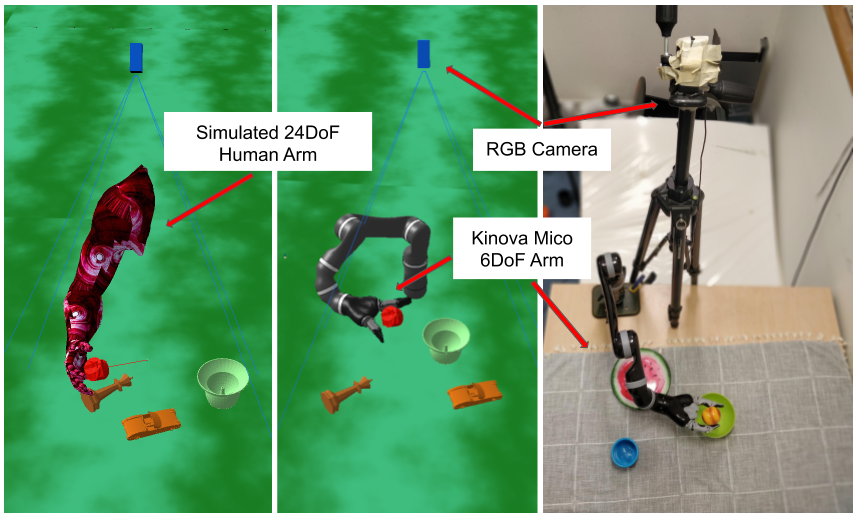}
    \caption{The simulation and real world setup. On the left, we see the 24 DoF arm, and in the centre we see the 6 DoF Kinova Mico arm; both have domain randomisation applied. On the right, we see real-world setup with the Kinova Mico. Observations come from an over-the-shoulder RGB camera in both simulation and the real world.}
    \label{fig:sim_and_real_scene}
\end{figure}

During dataset generation, we collect the image, the joint angles and the joint velocities at each timestep for both human arm and robot. To achieve sim-to-real transfer we perform domain randomisation. Specifically, we sample from a set of $5000$ textures and procedurally generated images (via Perlin noise), and apply them to all objects in the scene and to the human arm (an example can be seen in Figure \ref{fig:randomised_arm}). Additionally, we sample the position, the orientation and the size of the objects from a uniform distribution. The starting configuration of both the demonstrator and the agent, camera pose, light directions and lighting parameters are sampled from a normal distribution. A snapshot of the simulation and real-world setup can be seen in Figure \ref{fig:sim_and_real_scene}.

% Our dataset features a total of XXXXXX tasks, where each contains 15 simulated human demonstrations, and 15 simulated robot demonstrations. 

\subsection{Training} \label{training}

\begin{figure*}
    \centering
    \includegraphics[width=1.0\linewidth]{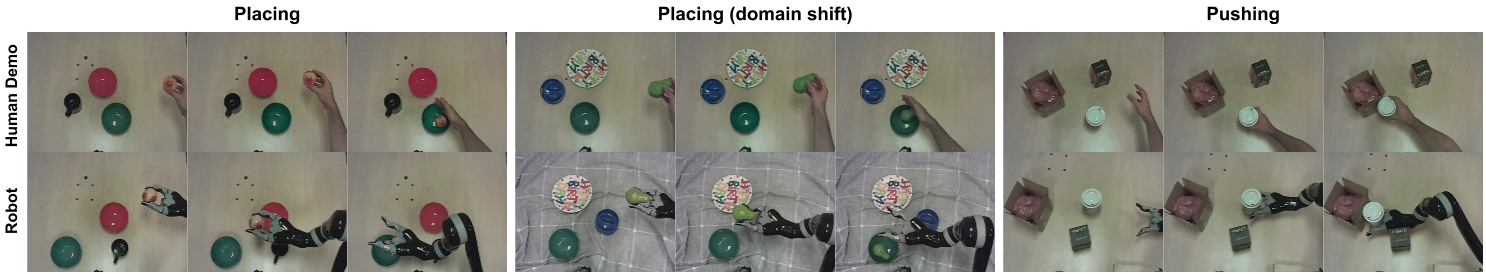}
    \caption{The three tasks that we evaluate our model on in the real world. Left: placing an object in one container with two distractor, same camera pose for human and robot and same background. Centre: the same placing task as on the left, but with a domain shift (cloth on the table) between the human demonstration and the robot execution. Right: pushing one object to a target with one distractor, same camera pose for human and robot and same background.}
    \label{fig:task_timeline}
\end{figure*}

\begin{figure}
\centering
\begin{subfigure}[]{0.4\linewidth}
   \includegraphics[width=0.95\linewidth]{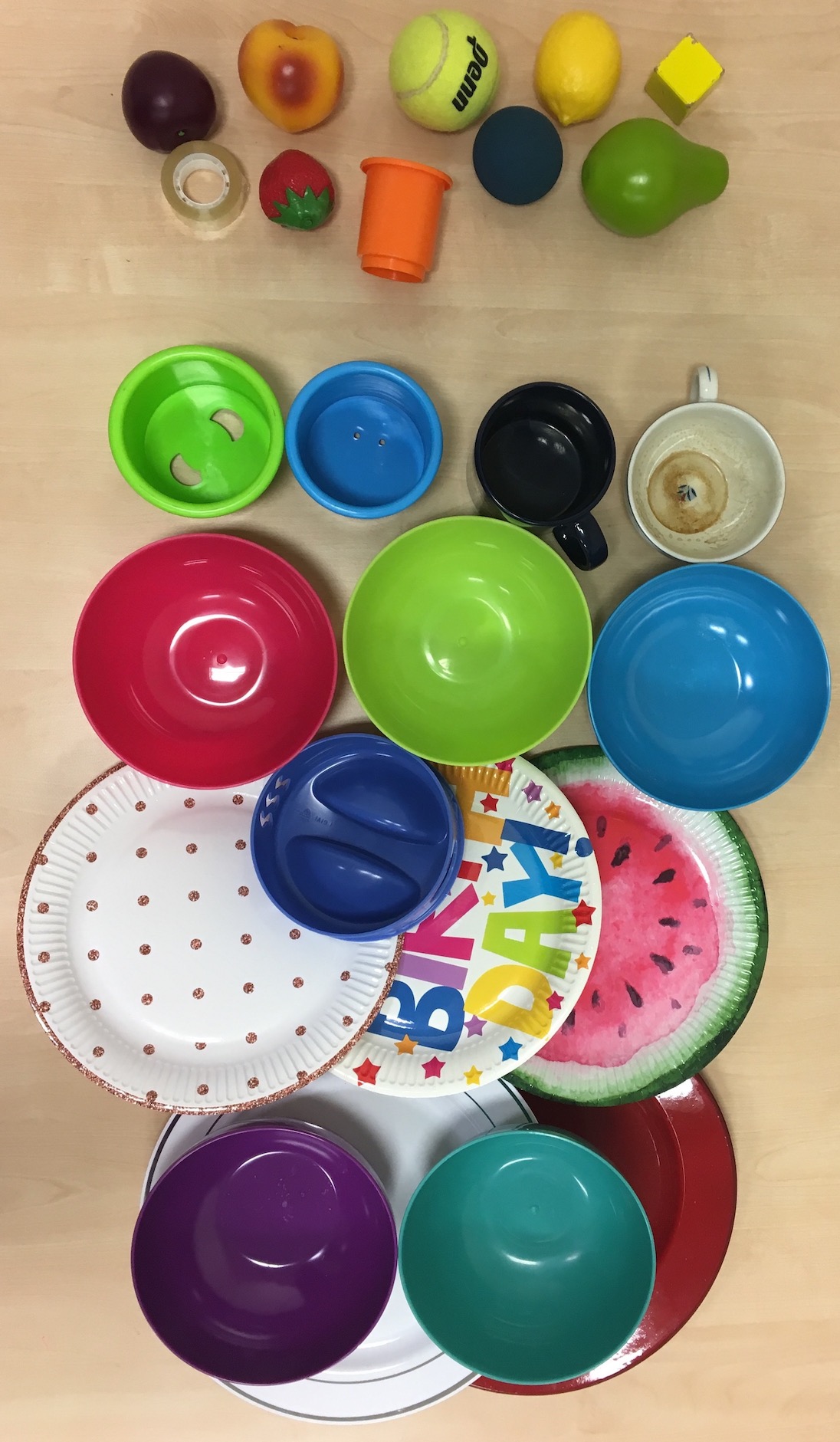}
   \caption{Placing}
   \label{fig:placingobjs} 
\end{subfigure}
\begin{subfigure}[]{0.51\linewidth}
   \includegraphics[width=0.95\linewidth]{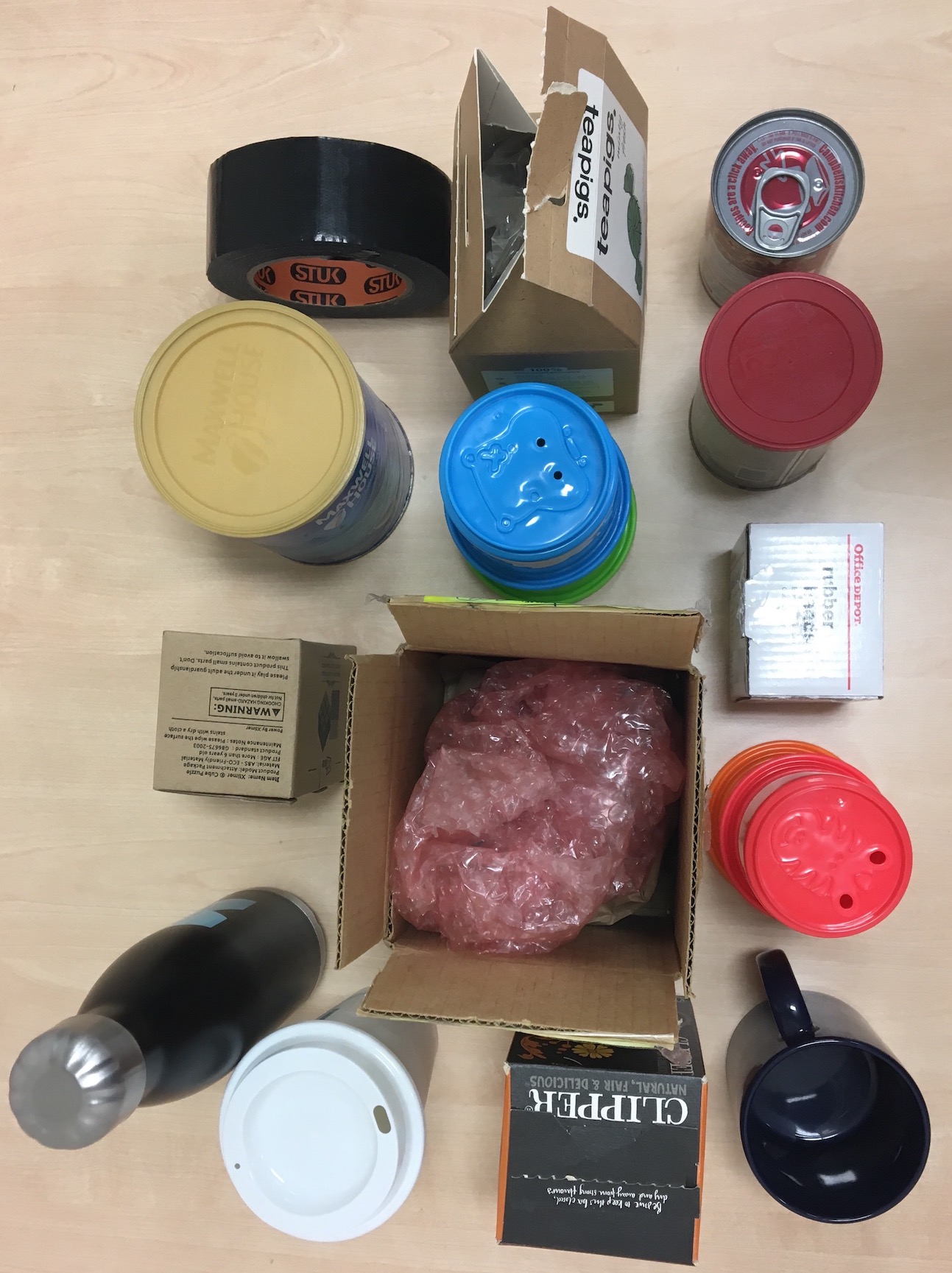}
   \caption{Pushing}
   \label{fig:pushingobjs}
\end{subfigure}
\caption{The real world test set for both the placing (a) and pushing (b) domain. For the placing domain (a), holding objects are on the top and placing objects (consisting of bowls, plates, cups, and pots) are on the bottom.}
\end{figure}

Our task-embedding network and control network use a convolutional neural network (CNN), which consists of $4$ convolution layers, each with $16$ filters of size $5\times5$, followed by $3$ fully-connected layers consisting of $200$ neurons. Each layer is followed by layer normalisation \cite{ba2016layer} and an $elu$ activation function \cite{clevert2015fast}, except for the final layer, where the output is linear for both the task-embedding and control network.

Input consists of a $125 \times 125$ RGB images and the robot proprioceptive data, including the joint angles. The proprioceptive data is concatenated to the features extracted from the CNN layers of the control network, before being sent through the fully-connected layers. The output of the embedding network (embedding size) is a vector of length $20$. The output of the control network corresponds to velocities applied to the $6$ joints of a Kinova Mico 6-DoF arm. During training, we set the margin to be $0.1$ for the embedding loss $\loss_{emb}$, and set both the support and query size to be $5$.

Optimisation was performed with Adam \cite{kingma2014adam} with a learning rate of $5 \times 10^{-4}$ and a batch size of $100$. Lambdas were set as follows: $\lambda_{emb} = 0.1$, $\lambda_{\humancollection_{ctr}}^{\support} = 1.0$, and $\lambda_{\robotcollection_{ctr}} = 1.0$.

\section{RESULTS}
In our experiments we try to answer the following questions: (1) Can TecNets learn the domain shift between a demonstrator and an agent? In other words, can our approach learn an embedding of a task given demonstrator examples, and also a mapping from the demonstrator domain to the agent domain for control? (2) Is it possible to learn a task from a real-world human demonstration when all the training is done is simulation? (3) How does our approach compare to another state-of-the-art one-shot human imitation learning method? We consider two experiments, placing and pushing, which were undertaken for DAML \cite{yu2018one} in order to compare our approach with their results. We run a set of experiments in both simulation and in the real world.

\subsection{Placing}
We begin by presenting our results for the placing experiment, both in simulation and in the real world: the goal is to place a hand-held object in a container, with other two containers in the scene acting as distractors. A trial is successful if the object lands inside the container. Our dataset features a total of 2280 tasks, where each contains 15 simulated human demonstrations, and 15 simulated robot demonstrations. For each task we sampled three objects from the MIL dataset \cite{finn2017one} of 105 training meshes, and used them as target containers and distractors; we randomised the scene as described in \ref{datacollection} and we trained the network in simulation with the parameters in \ref{training}.

We evaluated one-shot placing in simulation on 74 tasks, with 6 trials each, using the MIL test meshes: in every trial we randomise the position of the objects and of the camera, and we procedurally generate the hand-held object. We also performed evaluation for the same system in the real world (Figure \ref{fig:task_timeline} Left) on 18 tasks and 4 trials, using the containers and the held objects shown in Figure \ref{fig:placingobjs}, maintaining the same camera pose and background between demonstration and trial.

The results for the placing experiment are shown in Table \ref{table_place}. We find that the robot is able to learn from just one human demonstration of a previously unseen task, and can leverage the training with domain randomisation to bridge the reality gap, with comparable success rates to DAML. Additionally, we report the results of simulated placing evaluation for a network trained without the 
the control loss on the human examples support set $\loss_{\humancollection_{ctr}}^{\support}$. As it was previously outlined in \textit{James at al.} \cite{james2018task}, the inclusion of the support loss assists the network in learning the task and the mapping between domains.

We also report the results of a real world experiment with a dataset where we simply randomised the scene and made the held object float to the target bowl, without using our simulated human arm. The results show that without the simulated arm, the resulting real-world policy chooses a target at random; therefore the arm model is necessary to successfully learn to imitate from a single human demonstration.

\begin{table}[h]
\begin{center}
\begin{tabular}{|c|c|c|}
\hline
& Placing Experiment\\
\hline
\hline
DAML: Real World & 93.8\%\\
\hline
Ours: Real World & \textbf{88.9\%}\\
\hline
\hline
Ours: Sim & 94.1\%\\
\hline
Ours: Sim $\lambda_{ctr}^{\support} = 0$ & 78\%\\
\hline
Ours: Real World (No Sim Arm) & 39\% \\
\hline

\end{tabular}
\end{center}
\caption{One-shot success rate of the placing experiment, using novel objects. In simulation the scene is randomised for every trial. In the real-world the human demonstrations are taken from the perspective of the robot. We achieve comparable performance to DAML despite training on no real-world data.}
\label{table_place}
\end{table}

\subsection{Pushing}
In the pushing experiment the goal is to push an object against a target amid one distractor: a trial is successful if the object hits or falls within 5cm of the target. Our dataset features a total of 1620 tasks, with 15 domain randomised demonstrations for both robot and human, using objects from the MIL dataset.

We evaluated one-shot pushing in both simulation and real-world (Figure \ref{fig:task_timeline} Right), with the same number of trials as for the previous placing experiments. The objects used for the real-world experiment are shown in Figure \ref{fig:pushingobjs}. We report the results in Table \ref{table_push} together with the DAML results to show that our network trained in simulation has once again comparable performance to a model trained with real data.

\begin{table}[h]
\begin{center}
\begin{tabular}{|c|c|c|}
\hline
& Pushing Experiment\\
\hline
\hline
DAML: Real World & 88.9\%\\
\hline 
Ours: Real World & \textbf{84.7\%}\\
\hline
Ours: Sim & 87.6\%\\
\hline
\end{tabular}
\end{center}
\caption{One-shot success rate of the pushing experiment. In simulation the scene is randomised for every trial, whereas in the real-world the human demonstrations are taken from the perspective of the robot. As in the placing results, we achieve comparable performance to DAML despite training on no real-world data.}
\label{table_push}
\end{table}

\subsection{Large Domain Shift}
As a final experiment, we tested how resilient our model is against large domain shifts in the real world, expecting it to leverage the adaptability acquired from domain randomisation. We evaluated placing in the real world taking the human demonstrations with a cloth on the table, and then making the robot perform the same task with the table cloth removed, therefore with a substantial change of background (Figure \ref{fig:task_timeline} Centre): the model placed the held object correctly \textbf{87.5\% of the 72 trials}.

We have therefore shown that due to domain randomisation our performance does not degrade on large domain shifts, whereas for example in DAML the success rate under large change of scenes drops by up to 15\%. This showcases the benefits of leveraging large scale simulations for robotic learning.

%\begin{table}[h]
%\caption{Failure Modes}
%\label{table_example}
%\begin{center}
%\begin{tabular}{|c|c|c|c|}
%\hline
%Failure Analysis & Place & Push & Push novel bg \\
%\hline
%\# successes & 64 & & \\
%\hline
%\# failures from task identification & 5 & & \\
%\hline
%\# failures from control & 3 & & \\
%\hline
%\end{tabular}
%\end{center}
%\end{table}

\section{CONCLUSIONS}

We have presented an approach to the one-shot human imitation problem that leverages zero human interaction during training time. We achieve this by the combination of 2 methods. Firstly, we extending Task-Embedded Control Networks (\tecnets{}) \cite{james2018task} to infer control polices by embedding human demonstrations that can condition a control policy and achieve one-shot imitation learning. Secondly, and most importantly, we show that we are able to perform sim-to-real on humans which allows us to train our system with no real-world data. With this approach, we are able to achieve similar performance to a state-of-the-art alternative method that relies on thousands of training demonstrations collected in the real-world, whilst also remaining robust to visual domain-shifts, such as a substantially different backgrounds. For future work, we hope to further investigate the variety of human actions that can be transferred from simulation to reality. For example, in this work, we have shown that a human arm can be transferred, but would the same method work for demonstrations including the entire torso of a human? We hope this work provides the first step in answering this question.


%%%%%%%%%%%%%%%%%%%%%%%%%%%%%%%%%%%%%%%%%%%%%%%%%%%%%%%%%%%%%%%%%%%%%%%%%%%%%%%%

%%%%%%%%%%%%%%%%%%%%%%%%%%%%%%%%%%%%%%%%%%%%%%%%%%%%%%%%%%%%%%%%%%%%%%%%%%%%%%%%

\section*{ACKNOWLEDGMENT}
We thank Michael Bloesch, Ankur Handa, Sajad Saeedi, and Dan Lenton for insightful feedback on an early draft of this paper.



\bibliographystyle{IEEEtran}
\bibliography{ref}

\end{document}